\documentclass[10pt,twocolumn,letterpaper]{article}
\usepackage[utf8]{inputenc}
\usepackage{iccv}
\usepackage{times}
\usepackage{epsfig}
\usepackage{graphicx}
\usepackage{amsmath}
\usepackage{amssymb}

\usepackage{amsthm}

\usepackage{float}
\usepackage{adjustbox}
\usepackage{xcolor}
\usepackage{algorithm2e}

\DeclareMathOperator{\Tr}{tr}
\newtheorem*{theorem}{Theorem}

\def \calO {\mathcal{O}}
\newcommand{\psd}[2] {\mathcal{S}^+(#1,#2)}
\newcommand{\frmat}[2] {\mathbb{R}_*^{#1 \times #2}}
\def \Tu {\top \!}

\usepackage{hyperref}

\iccvfinalcopy 

\def\iccvPaperID{9} 

\ificcvfinal\fi

\begin{document}

\title{Fitting, Comparison, and Alignment of Trajectories on Positive Semi-Definite Matrices with Application to Action Recognition}

\author{\parbox{16cm}{\centering
    {\large Benjamin Szczapa$^1$, Mohamed Daoudi$^1$, Stefano Berretti$^2$, Alberto Del Bimbo$^2$, Pietro Pala$^2$ and Estelle Massart$^3$}\\
    {\normalsize
    $^1$ IMT Lille Douai, Univ. Lille, CNRS, UMR 9189 CRIStAL, F-59000 Lille, France\\
    $^2$ Department of Information Engineering, University of Florence, Italy \\
    $^3$ ICTEAM Institute, UCLouvain, Avenue Georges Lema\^{i}tre 4 bte L4.05.01, 1348 Louvain-la-Neuve, Belgium}}
}


\maketitle
\ificcvfinal\thispagestyle{empty}\fi

\begin{abstract}
In this paper, we tackle the problem of action recognition using body skeletons extracted from video sequences. Our approach lies in the continuity of recent works representing video frames by Gramian matrices that describe a trajectory on the Riemannian manifold of positive-semidefinite matrices of fixed rank. Compared to previous work, the manifold of fixed-rank positive-semidefinite matrices is endowed with a different metric, and we resort to different algorithms for the curve fitting and temporal alignment steps. We evaluated our approach on three publicly available datasets (UTKinect-Action3D, KTH-Action and UAV-Gesture). The results of the proposed approach are competitive with respect to state-of-the-art methods, while only involving body skeletons.
\end{abstract}

\section{Introduction}
In the last decades, automatic analysis of human motion has been an active research topic, with applications that have been exploited in a number of different contexts, including video surveillance, semantic annotation of videos, entertainment, human computer interaction and home care rehabilitation, to say a few.
Differences in body proportion (size, height, corpulence), body stiffness and training, influence the way different people perform an action. Even one same person is not able to perform the same action twice, exactly replicating the same sequence of body poses in space and time. This variability makes the task of human motion analysis very challenging.

For years, the approaches could be distinguished in two main classes: those operating on pixel values extracted from the RGB stream (either stacking groups of consecutive frames or extracting motion vectors) and those building upon the higher level representation of body skeletons. 
These latter approaches were supported by the diffusion of low-cost RGB-D cameras (such as the Microsoft Kinect) that can operate in real-time, while reliably extracting the 3D coordinates of body joints. 
More recently, deep CNN architectures have demonstrated real-time and accurate extraction of the coordinates of body joints from RGB streams~\cite{cao2017realtime}.
These advances make it possible to use a skeleton-based body representation in a much broader range of domains and operative contexts than before, being not limited by the short operative range of RGB-D sensors that typically operate indoor and in the range of a few meters. 
The design of the recognition/classification module on top of the body skeleton representation makes it possible to describe an action as a sequence of body poses, each one corresponding to a point in a feature space, whose dimension is proportional to the number of body joints. 
By exploiting the geometric properties of the manifold where these pose descriptors lie, it is possible to define a similarity metric that is invariant under translation, scaling, rotation and also under variations of the speed of execution of the action. 
Furthermore, the explicit representation of an action as a trajectory, \emph{i.e.}, a sequence of poses, on the manifold makes it possible to extract statistical summaries, such as mean and deviation from the mean, from a group of actions.
Through these summaries, one action can be better characterized for the purpose of detecting outliers corresponding to the anomalous execution of an action, that can be of particular relevance for action prediction. 
In fact, when analyzing the skeleton sequences, there are four main aspects to challenge: (1) A shape representation invariant to undesirable transformations; (2) A temporal modeling of landmark sequences; (3) A suitable rate-invariant distance between arbitrary sequences, and (4) A solution for temporal sequence classification.

\begin{figure*}[!ht]
  \centering
    \includegraphics[width=0.85\textwidth]{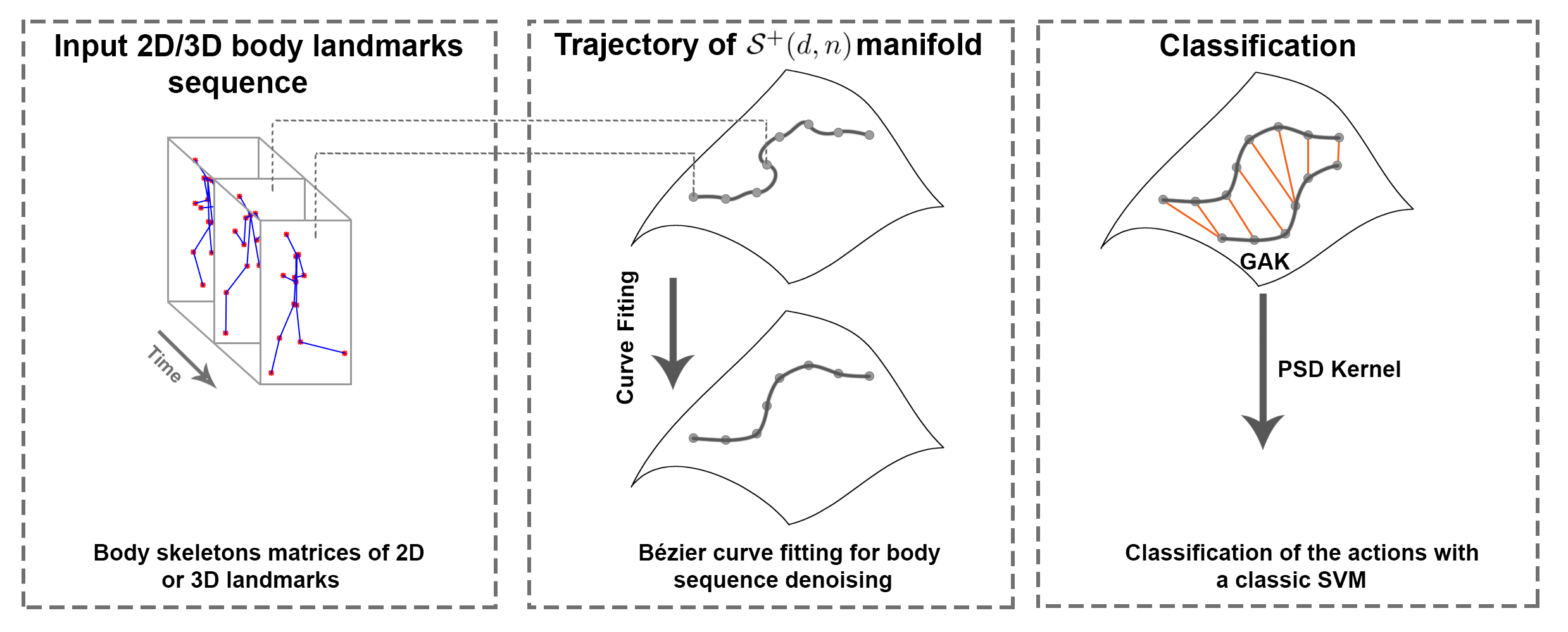}
    \caption{Overview of the proposed approach - After automatic body skeletons detection for each frame of a sequence, the Gram matrices are computed to build the trajectory on the $\mathcal{S}^+(d,n)$ manifold. We apply a curve fitting algorithm on the trajectory to smooth the curve and reduce noise. Global Alignment Kernel (GAK) is then used to align the trajectories on the manifold. Finally, we use the kernel generated from GAK with SVM to classify the actions.}
    \label{Fig:AppOverview}
\end{figure*}

This paper lies in the continuity of recent works that model the comparison and classification of temporal sequences of landmarks on the Riemannian manifold of positive-semidefinite matrices. Building on the work~\cite{KacemPAMI2019}, our approach involves four different steps: 1) We build a trajectory on the Riemannian manifold from the body skeletons; 2) We apply a curve fitting algorithm on the trajectories to denoise the data points; 3) We perform a temporal alignment using a Global Alignment Kernel, defining a positive-semidefinite kernel; 4) Finally, we use this kernel with a classic SVM to classify the actions. An overview of the full approach is given in Fig.~\ref{Fig:AppOverview}.



The novelties with respect to~\cite{KacemPAMI2019} are:
\begin{itemize}
\item The manifold of positive-semidefinite matrices is here endowed with a different metric;
\item A recent curve fitting method is used to smooth trajectories on the manifold;
\item We use Global Alignment Kernel for temporal alignment, instead of Dynamic Time Warping.
\end{itemize}

The rest of the paper is organized as follows: in Section~\ref{sect:2}, we summarize the closest works in the literature. In Section~\ref{sect:3}, we explain how we represent our data on the manifold of fixed-rank positive-semidefinite matrices, and present the Riemannian metric that we will use in this paper. In Section~\ref{sect:4}, we describe the curve fitting algorithm that we will use to denoise the trajectories. The curve alignment method and the classifier are presented in Section~\ref{sect5:classification}, while results and discussions are reported in Section~\ref{sect:exp-results}. Finally, in Section~\ref{sect:conclusion}, we conclude and draw some perspectives of the work.


\section{Related Work}\label{sect:2}
A detailed review of the many approaches to human action recognition and classification is out of the scope of this paper. The interested reader can refer to~\cite{kong-2018} for a detailed and updated survey.  
In the following, we focus on approaches that use body skeletons as inputs to the recognition/classification module.  

One of the first approaches to perform action recognition by the analysis of trajectories of tracked body interest points was presented in Matikainen \etal~\cite{Matikainen_2009_6466}. 
Despite the promising results obtained, the authors did not take into account the geometric information of the trajectories.
More recently, in the case of human skeleton in RGB-D images, Devanne \etal~\cite{Devanne2015Cybernetics}
proposed to formulate the action recognition task as the problem of computing a distance between trajectories generated by the joints moving during the action. An action is then interpreted as a normalized parameterized curve in $\mathbb{R}^N$. However, this approach does not take into account the relationship between the joints.
In the same direction, Su \etal~\cite{Su2014} proposed a metric that considers time-warping on a Riemannian manifold, thus allowing the registration of trajectories and the computation of statistics on the trajectories. Su \etal~\cite{Su_2014_CVPR} applied this framework to the problem of visual speech recognition. Similar ideas have been developed by Ben Amor \etal{}~\cite{amor2016action} on the Kendall's shape space with application to action recognition using rate-invariant analysis of skeletal shape trajectories.


Anirudh \etal~\cite{AnirudhTSS17} started from the framework of Transported Square-Root Velocity Fields (TSRVF), which has desirable properties including a rate-invariant metric and vector space representation. Based on this framework, they proposed to learn an embedding such that each action trajectory is mapped to a single point in a low-dimensional Euclidean space, and the trajectories that differ only in the temporal rate map to the same point. The TSRVF representation and accompanying statistical summaries of Riemannian trajectories are used to extend existing coding methods such as PCA, KSVD, and Label Consistent KSVD to Riemannian trajectories. In the experiments, it is demonstrated that such coding efficiently captures distinguishing features of the trajectories, enabling action recognition, stroke rehabilitation, visual speech recognition, clustering, and diverse sequence sampling.

In~\cite{vemulapalli2014human}, Vemulapalli \etal proposed a Lie group trajectory representation of the skeletal data on a product space of special Euclidean ($SE$) groups. For each frame, this representation is obtained by computing the Euclidean transformation matrices encoding rotations and translations between different joint pairs. The temporal evolution of these matrices is seen as a trajectory on $SE(3) \times \cdots \times SE(3)$ and mapped to the tangent space of a reference point. A one-versus-all SVM, combined with Dynamic Time Warping and Fourier Temporal Pyramid (FTP) is used for classification. One limitation of this method is that mapping trajectories to a common tangent space using the logarithm map could result in significant approximation errors. Aware of this limitation, in~\cite{vemulapalli2016rolling} the same authors proposed a mapping combining the usual logarithm map with a rolling map that guarantees a better flattening of trajectories on Lie groups.

More recently, Kacem \etal~\cite{KacemPAMI2019} proposed a geometric approach for modeling and classifying dynamic 2D and 3D landmark sequences based on Gramian matrices derived from the static landmarks. This results in an affine-invariant representation of the data. Since Gramian matrices are positive-semidefinite, the authors relies on the geometry of the manifold of fixed-rank positive-semidefinite matrices, and more specifically, to the metric investigated in~\cite{Bonnabel2009SIAM}. However, this metric is parametrized, and the parameter should ideally be learned from the data. In addition, this paper adopts Dynamic Time Warping for sequence alignment. The resulting distance does not generally lead to a positive-definite kernel for classification. 

All the approaches described above rely on the use of \textit{hand-crafted features} enabling representation of the action as a trajectory or point in some suitable manifold. Differently from these approaches, many neural network models have been proposed that rely on training for the extraction of \textit{deep learned features}.
Recurrent neural networks (RNNs) and particularly Long-Short-Term Memory Networks (LSTMs) have been used to perform action recognition by the analysis of sequences of skeleton poses~\cite{Zhu-2016}. However, these methods typically lose structural information when converting the skeleton data and joint connectivity into the vector-shaped input of the neural network. As an alternative, some approaches introduce Convolutional Neural Networks (CNNs)~\cite{Chao-2017} and Graph Convolutional Networks (GCNs)~\cite{Yan-2018,Li-2019} in the overall architecture so as to retain the structural information among joints of the skeleton.
Although these approaches result in state-of-the-art performance~\cite{Gao-2019} on public action recognition benchmarks, it is not possible to define a formal mathematical framework to compute a valid metric on the internal, learned feature representation so as to perform a statistical analysis of the learned actions. 

\section{Our Approach}\label{sect:3}
\subsection{Shape Representation}
%
To represent body movement dynamics, we rely on the time series made of the coordinates of the $n$ tracked body points (\emph{i.e.}, $p_1 = (x_1,y_1), \ldots, p_{n} = (x_{n},y_{n})$ in 2D, or $p_1 = (x_1,y_1,z_1), \ldots, p_{n} = (x_{n},y_{n},z_n)$ in 3D), during each video sequence.
Each video sequence is thus characterized by a set of landmark configurations $\{Z_0,\ldots, Z_{\tau}\}$, where $\tau$ is the number of frames of the video sequence, and where each configuration matrix $Z_{i}$ $(1 \leq i\leq \tau) \in  \mathrm{R}^{n \times d}$ encodes the position of the $n$ landmarks in $d$ dimensions (with $d = 2$ or $d = 3$). We aim to measure the dynamic changes of the curves made of the landmark configurations, remaining invariant to rotation and translation. 
Similarly as in~\cite{KacemPAMI2019}, this goal is achieved through a Gram matrix representation, where we compute the Gram matrices as:
\begin{equation}
\label{eq:gram}
G=Z Z^T\; .
\end{equation}

These Gram matrices are $n \times n$ positive-semidefinite matrices, of rank smaller than or equal to $d$ (always equal to $d$ in the datasets considered). Conveniently for us, the Riemannian geometry of the space ${\mathcal S}^+(d,n)$ of $n \times n$ positive-semidefinite matrices of rank $d$ has been studied in~\cite{Bonnabel2009SIAM,journee2010low,vandereycken2009embedded,vandereycken2013riemannian,Massart2018, massart2019curvature}, and used in, \emph{e.g.},~\cite{faraki2016image,meyer2011regression,gousenbourger2017piecewise, massart2019interpolation}.

A classical approach in the design of algorithms on manifolds consists in resorting to first order local approximations on the manifold, called tangent spaces. This requires two tools: the Riemannian exponential (that allows us to map tangent vectors from the tangent space to the manifold), and the Riemannian logarithm (mapping points from the manifold to the tangent space).

In~\cite{KacemPAMI2019}, the manifold $\psd{d}{n}$ is identified to the quotient manifold $(\mathrm{St}(d,n) \times \mathcal{P}_d)/\calO_d$, where $\mathrm{St}(d,n) := \{  Y \in \mathbb{R}^{n \times d} \rvert Y^T Y = I_d \}$ is the Stiefel manifold, $\mathcal{P}_d$ is the manifold of $d \times d$ positive-definite matrices, and $\calO_d$ is the orthogonal group in dimension $d$. We consider here another representation of the manifold $\psd{d}{n}$, that will result in different expressions for the distance between two points, the Riemannian exponential and logarithm.

\subsection{The Quotient Manifold $\frmat{n}{d} / \calO_d$}\label{metrics}

We consider here the identification of $\psd{d}{n}$ to the quotient manifold $\frmat{n}{d} / \calO_d$, where $\frmat{n}{d}$ is the set of full-rank $n \times d$ matrices. This geometry has been studied in~\cite{journee2010low,Massart2018,massart2019curvature}. 

The identification of $\psd{d}{n}$ with the quotient $\frmat{n}{d} / \calO_d$ comes from the following observation. Any PSD matrix $G \in \psd{d}{n}$ can be factorized as $G = ZZ^T$, with $Z \in \frmat{n}{d}$. However, this factorization is not unique, as any matrix $\tilde Z := ZQ$, with $Q \in \calO_d$, satisfies $\tilde Z \tilde Z^T = Z QQ^T Z^T = G$. The two points $Z$ and $\tilde Z$ are thus \emph{equivalent} with respect to this factorization, and the set of equivalent points
\[
Z \calO_d := \{  ZQ | Q \in \calO_d \}, 
\]

\noindent 
is called the equivalence class associated to $G$. The quotient manifold $\frmat{n}{d} / \calO_d$ is defined as the set of equivalence classes. The mapping $\pi : \frmat{n}{d} \to \frmat{n}{d} / \calO_d$, mapping points to their equivalence class, induces a Riemannian metric on the quotient manifold from the Euclidean metric in $\frmat{n}{d}$. This metric results in the following distance between PSD matrices:
\begin{equation}
\label{eq:PSD-norm}
    d(G_i, G_j) = \left[  \mathrm{tr}(G_i) + \mathrm{tr}(G_j) - 2 \mathrm{tr}\left( \left( G_i^{\frac{1}{2}}  G_j G_i^{\frac{1}{2}} \right)^{\frac{1}{2}} \right) \right]^{\frac{1}{2}}.
\end{equation}

This distance can be expressed in terms of the landmark variables $Z_i, Z_j \in \frmat{n}{d}$ as follows:
\begin{equation}
\label{eq:frob-norm-new-metrix}
d(G_i, G_j) = \min_{Q \in \calO_d} \lVert Z_jQ-Z_i\rVert_F. 
\end{equation}

\noindent
The optimal solution is $Q^* := VU^\Tu$, where $Z_i^\Tu Z_j = U \Sigma V^\Tu$ is a singular value decomposition.



As stated by the next theorem, when $d = 2$, the distance can also be formulated as follows: 

\begin{theorem}
\label{theo1}
Let $G_i, G_j \in \psd{2}{n}$ be two Gram matrices, obtained from landmark matrices $Z_i, Z_j \in \mathbb{R}^{n \times 2}$. The Riemannian distance~\eqref{eq:PSD-norm} can be expressed as:
%
\begin{equation} 
\label{eq:distp2}
\resizebox{1\hsize}{!}{$d(G_i, G_j)= \left( \Tr(G_i) - 2\sqrt{(a + d)^2 + (c - b)^2} + \Tr(G_j) \right)^{\frac{1}{2}}$}, 
\end{equation}
\end{theorem}

\noindent 
where $Z_j^TZ_i$ = $\left( \begin{array}{cc} a & b \\ c & d \end{array} \right)$.
\begin{proof}
See Appendix.
\end{proof}

Expressions for the Riemannian exponential and logarithm are given in~\cite{Massart2018}. We used the implementations provided in the Manopt toolbox~\cite{manopt}.
\section{Trajectory Modeling}\label{sect:4}
The dynamic changes of body joints movement are characterized by trajectories on the Riemannian manifold of positive-semidefinite matrices of fixed rank (see Fig.~\ref{Fig:AppOverview}). More specifically, we fit a curve $\beta_G$ to a sequence  of landmark configurations $\{Z_0, \ldots, Z_\tau\}$ represented by their corresponding Gram matrices $\{G_0,\ldots,G_\tau\}$ in $\mathcal{S}^+(d,n)$. This curve will enable us to model the spatio-temporal evolution of the elements on $\mathcal{S}^+(d,n)$. 

Modeling a sequence of landmarks as a piecewise-geodesic curve on $\mathcal{S}^+(d,n)$ showed very promising results when the data are well acquired, \emph{i.e.}, without tracking errors or missing data, see~\cite{KacemPAMI2019, KacemICCV17, OtberdoutBMVC2018}. To account for both missing data and tracking errors, we rely on a more recent curve fitting algorithm: fitting by composite cubic blended curves, proposed in~\cite[\S 5]{Gousenbourger2018}. Specifically, given a set of points $G_0,\ldots,G_\tau \in \mathcal{S}^+(d,n)$ associated to times $t_0, \dots, t_\tau$, with $t_i := i$, the curve $\beta_G$, defined on the interval $[0,\tau]$, is defined as:
\[ 
\beta_G(t) := \gamma_i(t-i), \qquad t \in [i, i+1], 
\]

\noindent
where each curve $\gamma_i$ is obtained by blending together fitting cubic B\'ezier curves computed on the tangent spaces based on the data points $d_i$ and $d_{i+1}$. 

These fitting cubic B\'ezier curves depend on a parameter $\lambda$, allowing us to balance two objectives: proximity to the data points at the associated time instants, and regularity of the curve (measured in terms of mean square acceleration). A high value of $\lambda$ results in a curve with possibly high acceleration, but that will almost interpolate the data, while taking $\lambda \rightarrow 0$ will result in a geodesic. The interested reader can refer to~\cite[\S 5]{Gousenbourger2018} for more information about the curve fitting procedure. 



\section{Classification}\label{sect5:classification}
Now that we have defined how to represent a sequence and how to compare two distinctive landmark configurations, we present in this section how we compare two landmark sequences and how to classify the actions performed in these same sequences.

\subsection{Global Alignment}\label{global-alignment-kernel}
As we described in Section~\ref{sect:4}, we represent a sequence as a trajectory of Gram matrices in $\psd{d}{n}$. The sequences represented in this manifold can be of different length as the execution rate of the actions can vary from one person to another, meaning that we can not effectively compare them. A common method to do so is to use Dynamic Time Warping (DTW) as proposed in several works~\cite{amor2016action, KacemPAMI2019, gritai2009matching}. However, DTW does not define a proper metric and can not be used to derive a valid positive-definite kernel for the classification phase. To address the problem of non positive definiteness of the kernel defined by DTW, Cuturi \etal~\cite{CuturiVBM07} proposed the Global Alignment Kernel (GAK), which allows us to derive a valid positive-definite kernel when aligning two time series. More recently Otberdout \etal~\cite{Otberdout2019} have proposed to classifiy deep trajectories in SPD manifold using GAK. The generated kernel can be used directly with Support Vector Machine (SVM) for the classification phase, whereas it is not the case with kernels generated with DTW. In fact, the kernels built with DTW do not show favorable positive definiteness properties as they rely on the computation of an optimum rather than the construction of a feature map. Note that the computation of the kernels with GAK can be done in quadratic complexity, similarly to naive implementation of DTW. The next paragraph describes how to compute the similarity score between two sequences, using this Global Alignment Kernel.



Let us now consider $Z^1= \{Z^1_0,\cdots, Z^1_{\tau_1}\}$ and $Z^2=\{Z^2_0,\cdots, Z^2_{\tau_2}\}$, two sequences of landmark configuration matrices. Given a metric to compute the distance between two elements of each sequence, we propose to compute the matrix $D$ of size $\tau_1 \times \tau_2$, where each $D(i,j)$ is the distance between two elements of the sequences, with $1 \leq i \leq \tau_1$ and $1 \leq j \leq \tau_2$.
\begin{equation}
\label{eq:matrix-distance}
    D(i,j) = d(Z^1_i, Z^2_j) .
\end{equation}

The kernel $\tilde{k}$ can now be computed using the halved Gaussian Kernel on this same matrix $D$. Therefore, the kernel $\tilde{k}$ can be defined as:
\begin{equation}
\label{eq:k-tilde}
    \tilde{k}(i,j) = \frac{1}{2} * exp\left(-\frac{D(i,j)}{\sigma^2}\right).
\end{equation}
 As reported in~\cite{CuturiVBM07}, we can redefine our kernel such as:
\begin{equation}
\label{eq:kernel-PSD}
    k(i,j) = \frac{\tilde{k}(i,j)}{(1-\tilde{k}(i,j))} .
\end{equation}
This strategy assures us that the kernel is positive semi-definite and can be used in its own. Finally, we can compute the similarity score between the two sequences $Z^1$ and $Z^2$. Remember that this computation is performed in quadratic complexity, like DTW. To do so, we define a new matrix $M$ that will contain the path to the similarity between our two sequences. We define $M$ as a zeros matrix of size $(\tau_1+1) \times (\tau_2+1)$ and $M_{0,0} = 1$. Computing the terms of $M$ is done using Theorem 2 in~\cite[\S 2.3]{CuturiVBM07}:
%
%
\begin{equation}
\label{eq:matrix-similarity}
    M_{i,j} = (M_{i,j-1} + M_{i-1,j-1} + M_{i-1,j})*k(i,j) .
\end{equation}

The similarity score we seek is the value at $M_{(\tau_1+1),(\tau_2+1)}$. Algorithm~\ref{algo:similarity} describes all the steps to get the similarity score.

Finally, we build a new matrix $K$ of size $n_{seq} \times n_{seq}$, where $n_{seq}$ is the number of sequences in the dataset we test. This matrix is symmetric and contains all the similarity scores between all the sequences of the dataset and it is used as the kernel for the classification phase with SVM. As this matrix is built with values computed from positive semi-definite kernel, it is a positive semi-definite matrix itself.

\begin{algorithm}
\caption{Computing the similarity score between two sequences using Global Alignment Kernel~\cite{CuturiVBM07}
\label{algo:similarity}}
\DontPrintSemicolon
\SetKwInOut{Input}{input}\SetKwInOut{Output}{output}
\Input{Two sequences of landmark configurations $Z^1= \{Z^1_0,\cdots, Z^1_{\tau_1}\}$, where $Z^1_{0 \leq i \leq \tau_1}$ and $Z^2=\{Z^2_0,\cdots, Z^2_{\tau_2}\}$, where $Z^2_{0 \leq j \leq \tau_2}$.}
\Output{The similarity score between two sequences $Z^1, Z^2$}
$\tilde{k} \longleftarrow \frac{1}{2}*exp\left(-\frac{D(Z^1, Z^2)}{\sigma^2} \right)$ Equations ~\eqref{eq:matrix-distance} and \eqref{eq:k-tilde} \\

\For{$i \leftarrow 0$ \KwTo $\tau_1$}{
    \For{$j \leftarrow 0$ \KwTo $\tau_2$}{
        $k(i,j) \longleftarrow \frac{\tilde{k}(i,j)}{(1 - \tilde{k}(i,j))}$ Equation~\eqref{eq:kernel-PSD}
    }
}

$M \longleftarrow zeros(\tau_1+1, \tau_2+1)$ \\
$M_{0,0} \longleftarrow 1$ \\

\For{$i \leftarrow 1$ \KwTo $\tau_1+1$}{
    \For{$j \leftarrow 1$ \KwTo $\tau_2+1$}{
        $M_{i,j} \longleftarrow (M_{i,j-1} + M_{i-1,j-1} + M_{i-1,j})*k(i,j)$ See Equation~\eqref{eq:matrix-similarity}
    }
}
$similarity \longleftarrow M_{\tau_1+1, \tau_2+1}$

\Return similarity, the similarity score between $Z^1$ and $Z^2$
\end{algorithm}

\subsection{Classification with SVM}
Our trajectory representation reduces the problem of landmark sequence classification to that of trajectory classification in $\mathcal{S}^+(d,n)$. Given that GAK provides a valid PSD kernel as demonstrated by Cuturi \etal{}~\cite{CuturiVBM07}, and given that our local kernel $K$ satisfies this condition as discussed before, we use the standard SVM with the $K$ kernel that represents the matrix containing the similarity scores between all the sequences of a dataset to classify the aligned trajectories with global alignment on $\mathcal{S}^+(d,n)$.

By contrast, DTW may define a non positive definite kernel. Hence, we adopt the pairwise proximity function SVM (ppfSVM), which assumes that instead of a valid kernel function, all that is available is a proximity function without any restriction. 
That is, let us consider $\mathcal{T} = \{\beta_G~:[0,1]\rightarrow \mathcal{S}^+(d,n)\}$, the set of time-parameterized trajectories of the underlying manifold. Like in~\cite[\S 4.1]{KacemPAMI2019}, we define a matrix $D_{dtw}$ containing the similarity measure between two trajectories aligned with DTW.
In that case, given $m$ trajectories in $\mathcal{T}$, the proximity function $\mathcal{P} : \mathcal{T} \times \mathcal{T} \rightarrow \mathbb{R}^+$ between two trajectories $Z^{1}$ and $Z^{2}$ is defined by,
\begin{equation}
\mathcal{P}(Z^{1}, Z^{2})=D_{dtw}(Z^{1}, Z^{2}) \; .
\end{equation}

\noindent
Using this proximity function, the main idea of ppfSVM is to represent each training example $Z$ with a vector $[\mathcal{P}(Z,Z^1), \dots , \mathcal{P}(Z,Z^m)]^T$. The set of trajectories can be represented by a $m \times m$ matrix $P$, where $P(i,j) = \mathcal{P}(Z^{1}, Z^{2})$, with $1 \leq i,j \leq m$. From this matrix $P$ we can use a classical linear SVM.

\section{Experimental Results}
\label{sect:exp-results}
To validate the proposed approach, we have conducted experiments on three publicly available datasets with 3D and 2D actions: UTKinect-Action3D, KTH-Action and UAV-Gesture. Our experiments followed the experimental settings commonly used for these datasets.

\subsection{UTKinect-Action3D Dataset}
The UTKinect-Action3D dataset~\cite{xia2012view} is a widely used dataset for 3D action recognition. It contains 199 sequences, consisting of 10 actions, namely \textit{walk, sit down, stand up, pick up, carry, throw, push, pull, wave hands} and \textit{clap hands} performed by 10 different subjects. The videos and the skeletons were captured with a Microsoft Kinect and the skeletons are composed of 20 body joints. In our approach, we use the available skeletal joint locations, where each body joint is defined with its $x$, $y$ and $z$ coordinates. Following the same experimental settings of~\cite{TanfousDA18, LiuWDAK18, KeBASB18}, we performed the Leave-One-Out cross validation, meaning that we used one sequence for testing and the rest for training. 
Our experimental results are summarized in Table~\ref{results-our-utk}. In particular, the columns are as follows: \emph{Curve Fitting} indicates if we performed the curve fitting algorithm described in Section~\ref{sect:4}; \emph{Lambda} indicates the value of the lambda parameter in curve fitting; 
\emph{Alignment Method} indicates if we used the standard DTW to align sequences or GAK as described in Section~\ref{global-alignment-kernel}; \emph{Sigma} indicates the value of the sigma parameter for the Gaussian Kernel when using GAK; and \emph{Results} indicates our scores.


\begin{table}[!ht]
\centering
\caption{Our results on the UTKinect-Action3D dataset}
\begin{adjustbox}{width=0.47\textwidth}
\begin{tabular}{|c|c|c|c|c|}
\hline
\textbf{Curve Fitting} & \textbf{Lambda} & \textbf{Alignment Method} & \textbf{Sigma} & \textbf{Results} \\ \hline
Yes & 0.5 & DTW & - & 97\% \\
No & - & DTW & - & 97.49\% \\ \hline
Yes & 0.5 & GAK & 0.3 & 97.49\% \\
No & - & GAK & 0.3 & \textbf{97.99\%} \\
Yes & 0.5 & GAK & 0.5 & \textbf{97.99\%} \\ \hline
\end{tabular}
\end{adjustbox}
\label{results-our-utk}
\end{table}

The best accuracy that we obtained on this dataset is 97.99\%. Overall, we can say that the application of curve fitting does not increase our results. Our assumption is that the data in this dataset are very clean, and we can loose some information with the application of smoothing on clean data. Note that we obtained better results when using the Global Alignment Kernel rather that DTW. 

\begin{table}[H]
\centering
\caption{Comparison of our approach with state-of-the-art results for the UTKinect-Action3D dataset. *: Deep Learning approach}
\begin{adjustbox}{width=0.38\textwidth}
\begin{tabular}{|l|c|c|}
\hline
 & \multicolumn{2}{c|}{\textbf{Protocol}} \\ 
 \hline
\textbf{Methods} & \textbf{H-H} & \textbf{LOOCV} \\ \hline
Trajectory on ${\mathcal S}^+(d,n)$~\cite{KacemPAMI2019} (2019) & - & 96,48\% \\
SCK+DCK~\cite{KoniuszCP16} (2016) & 98.2\% & - \\
Bi-LSTM~\cite{TanfousDA18} (2018)* & - & 98.49\% \\
LM$^3$TL~\cite{YangDTZLG17} (2017) & - & 98.8\% \\
GCA-LSTM~\cite{LiuWDAK18} (2018)* & - & 99\% \\
MTCNN~\cite{KeBASB18} (2018)* & - & 99\% \\
Hankel \& Gram matrices~\cite{ZhangWGSC16} (2016) & - & \textbf{100\%} \\ 
\hline
Ours & - & 97.99\% \\ 
\hline
\end{tabular}
\end{adjustbox}
\label{sota-utk}
\end{table}

In Table~\ref{sota-utk}, we compare our method with recent state-of-the-art results. Overall, our approach achieves competitive results with respect to most recent approaches. We directly compare our results with~\cite{KacemPAMI2019} as we work on the same geometric space of ${\mathcal S}^+(d,n)$ manifold. The main differences between our method and the method in~\cite{KacemPAMI2019} is the use of a different metric and of the Global Alignment Kernel instead of DTW. Our metric is simpler that the metric in~\cite{KacemPAMI2019}, as we do not have to estimate the parameter $k$ used in Eq.~(7) in~\cite{KacemPAMI2019} for distance computation. Furthermore, the $k$ parameter in~\cite{KacemPAMI2019} is more of a constraint as they have to determine its best value for each dataset they test. The use of GAK is also an advantage for us as it defines a positive semi-definite kernel, which is not the case for DTW allowing us to use a classic SVM instead of ppfSVM.

\begin{figure}[H]
    \centering
    \includegraphics[width=0.47\textwidth]{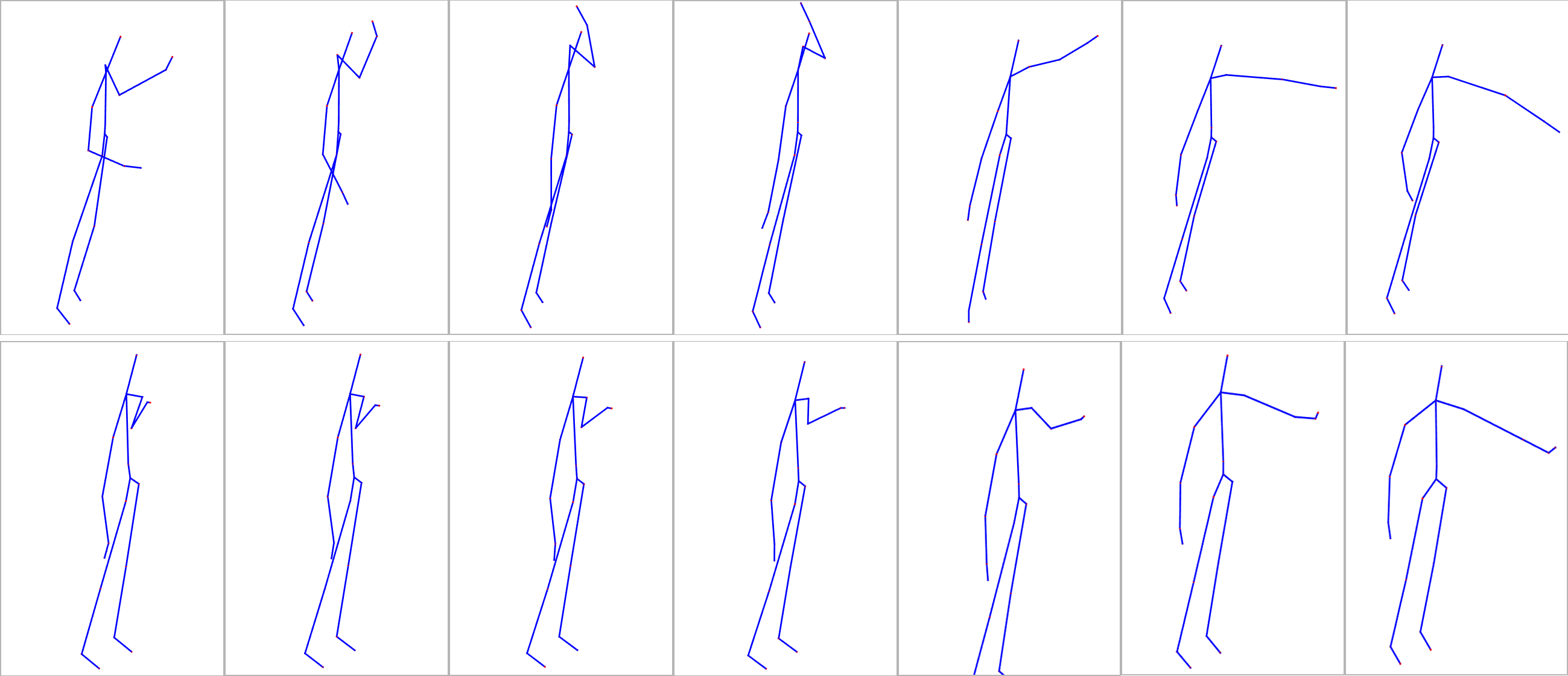}
    \caption{Comparison of two sequences that are confused in UTKinect-Action3D dataset (top: \textit{Throw} action, bottom: \textit{Push} action)}
    \label{qualitative-utk}
\end{figure}

The Figure~\ref{qualitative-utk} presents two sequences that are confused, leading to a misclassification for one of them. In that case, the top action (\emph{i.e.} Throw) is misclassified as the bottom action (\emph{i.e.} Push). One of the reasons can be the position of the arm at the end of the action, which is the same in the two sequences. The \textit{Throw} action is the most confused action in the dataset.

\subsection{KTH-Action Dataset}
\begin{figure}[H]
    \centering
    \includegraphics[width=0.22\textwidth]{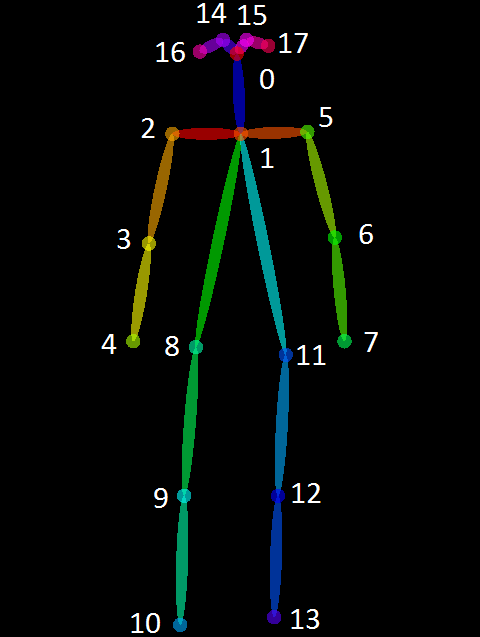}
    \caption{Skeleton with the COCO format.}
    \label{coco-skeleton}
\end{figure}

The KTH-Action dataset~\cite{kth} is a 2D action recognition dataset. It consists of six actions, namely \textit{boxing, handclapping, handwaving, jogging, running} and \textit{walking} performed by 25 subjects in four different conditions, which are outdoor, outdoor with scale variations, outdoor with different clothes and indoor. The sequences were acquired with a static camera at a frame rate of 25 fps and a resolution of $160 \times 120$ pixels. The dataset contains a total of 599 clips, with 100 clips per actions (1 clip is missing for one action). As the sequences in the dataset are 2D videos, we have to extract the skeletons of the subjects performing the actions. To do so, we used the OpenPose framework~\cite{openpose} to extract the skeletons in the COCO format, with 18 body joints. Note that we clean the landmark sequences by removing the frames where the body joints where not effectively estimated. Keeping all the frames leads to worst results due to misdetected joints, meaning that we do not need all the frames available to recognize an action. Figure~\ref{coco-skeleton} shows the configuration of the body joints that we analyzed. For this dataset, we followed the Leave-One-Actor-Out cross validation protocol, meaning that we use one subject for testing and the rest for training. Table~\ref{results-ours-kth} summarizes our experimental results on this dataset. 

\begin{table}[H]
\centering
\caption{Our experimental results on the KTH-Action dataset}
\begin{adjustbox}{width=0.47\textwidth}
\begin{tabular}{|c|c|c|c|c|}
\hline
\textbf{Curve Fitting} & \textbf{Lambda} & \textbf{Alignment Method} & \textbf{Sigma} & \textbf{Results} \\ \hline
No & - & DTW & - & 94.49\% \\
Yes & 10 & DTW & - & 94.66\% \\ \hline 

No & - & GAK & 0.2 & 95.16\% \\
Yes & 10 & GAK & 0.2 & \textbf{96.16\%} \\ \hline
\end{tabular}
\end{adjustbox}
\label{results-ours-kth}
\end{table}


Here, again, we obtained better results when using the GAK, demonstrating superior performance over DTW. The results reported with DTW are the best accuracy over all the configurations we tested. Unlike the data in UTKinect-Action3D dataset, the data in KTH-Action are 2D and low resolution videos, with presence of noise in the background, leading to noisy skeleton data after extraction. In this regard, the application of the curve fitting algorithm improves our results by 1\%. We compare our approach with the state-of-the-art in Table~\ref{sota-kth}. Overall, our method achieves competitive results with recent approaches, while only using skeletal data.

\begin{table}[H]
\centering
\caption{Comparison of our approach with state-of-the-art results for the KTH-Action dataset. *: Deep Learning approach}
\begin{adjustbox}{width=0.41\textwidth}
\begin{tabular}{|l|c|c|c|}
\hline
\textbf{Methods} & \textbf{Input data} & \textbf{Protocol} & \textbf{Accuracy} \\ 
\hline
Schüldt \etal~\cite{kth} (2004) & RGB & Split & 71.7\% \\
Liu \etal~\cite{LiuLS09} (2009) & RGB & LOAO & 93.8\% \\
Yoon \etal\cite{YoonK10} (2010) & Skeleton & - & 89\% \\
Raptis \& Soatto~\cite{RaptisS10} (2010) & RGB & LOAO & 94.5\% \\
Wang \etal~\cite{WangKSL11} (2011) & RGB & Split & 94.2\% \\
Gilbert \etal~\cite{GilbertIB11} (2011) & RGB & LOAO & 95.7\% \\
Jiang \etal~\cite{JiangLD12} (2012) & RGB & LOAO & 95.77\% \\
Vrigkas \etal~\cite{VrigkasKNK14} (2014) & RGB & LOAO & \textbf{98.3\%} \\
Veeriah \etal~\cite{VeeriahZQ15} (2015)* & RGB & Split & 93.96\% \\
Liu \etal~\cite{LiuSLL16} (2016) & RGB & Split & 95\% \\
Almeida \etal~\cite{AlmeidaPG17} (2017) & RGB & LOAO & 98\%  \\ 
\hline
Our & Skeleton & LOAO & 96.16\% \\ 
\hline
\end{tabular}
\end{adjustbox}
\label{sota-kth}
\end{table}

\subsection{UAV-Gesture Dataset}
The UAV-Gesture dataset~\cite{UAV-Gesture} is a 2D videos dataset, consisting of 13 actions corresponding to UAV (\emph{i.e.}, Unmanned Aerial Vehicles) gesture signals. These actions are \textit{All Clear, Have Command, Hover, Land, Landing Direction, Move Ahead, Move Downward, Move To Left, Move To Right, Move Upward, Not Clear, Slow Down} and \textit{Wave Off}. The actions are performed by 11 different subjects in an outdoor scenario with slight camera movements. The dataset contains 119 high-quality clips consisting of 37151 frames. As reported in~\cite{UAV-Gesture}, this dataset is not primarily designed for action recognition, but it can be used for this specific task. The skeletons are available with the dataset and the OpenPose framework was also used to extract them in the COCO format. Table~\ref{results-ours-uav} compares our results with the baseline experiment reported in~\cite{UAV-Gesture}.

\begin{table}[H]
\centering
\caption{Comparison of our approach with the baseline on the UAV-Gesture dataset. *: Deep Learning approach}
\begin{adjustbox}{width=0.478\textwidth}
\begin{tabular}{|c|c|c|c|c|}
\hline
\textbf{Method} & \textbf{Curve Fitting} & \textbf{Lambda} & \textbf{Alignment Method} & \textbf{Results} \\ \hline
P-CNN~\cite{UAV-Gesture} (2018)* & - & - & - & 91.9\% \\ \hline
Ours & No & - & GAK & 91.6\% \\
Ours & Yes & 10 & GAK & \textbf{92.44\%} \\ \hline
\end{tabular}
\end{adjustbox}
\label{results-ours-uav}
\end{table}

This is a very recent dataset and its principal interest does not rely on action recognition, meaning a lack of results to compare our results with. However, the authors have tested their dataset for the case of action recognition based on skeletons with Pose-Based Convolutional Neural Network (P-CNN) descriptors, that gives us a baseline to compare our results. 
The baseline achieves an accuracy of 91.9\% with a Deep Learning based approach, whereas our approach achieves an accuracy of 92.44\%, outperforming the state-of-the-art results when applying curve fitting and the GAK alignment method.

\begin{figure}[H]
    \centering
    \includegraphics[width=0.47\textwidth]{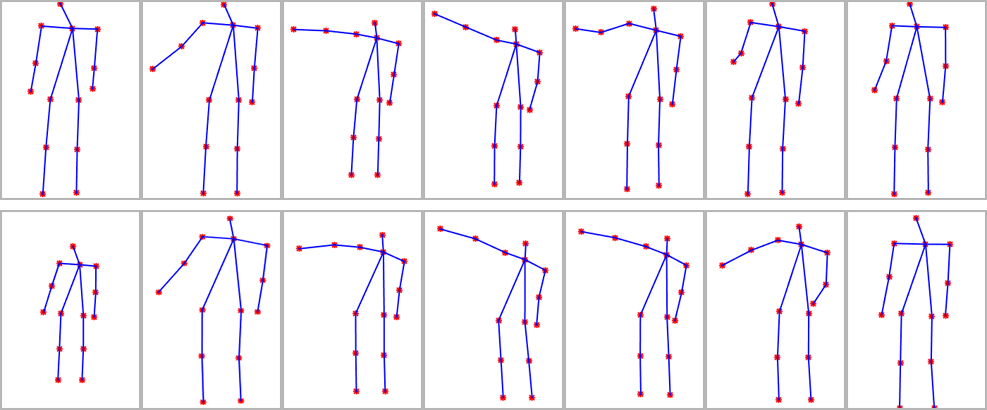}
    \caption{Comparison of two sequences that are confused in UAV-Gesture dataset (top: \textit{All Clear} action, bottom: \textit{Not Clear} action)}
    \label{qualitative-uav}
\end{figure}

The Figure~\ref{qualitative-uav} presents two very similar actions in the dataset, that is \textit{All Clear} and \textit{Not Clear}. The only big difference between these two actions is the orientation of the hand on the raised arm. 
This information is not captured when only retrieving the body skeleton from the OpenPose framework, even if it is possible to retrieve the hand skeleton. Note that with our method, the two actions are only confused three time on a total of 22 sequences, meaning that our method is capable of differentiate minimal changes in the actions.

\subsection{Computation Time Comparison}
In this analysis, we have computed the time for each step of our approach. Applying the curve fitting algorithm can be resource demanding on some manifolds, as it requires successive computations of Riemannian exponentials and logarithms (see~\cite{Gousenbourger2018} for more information on the computational cost of the method). The alignment method can also be resource demanding, regarding the size of the trajectory. With all these parameters in mind, we propose to compute the time that our method takes to compute specific tasks from pose extraction to action classification. The tests were conducted on a laptop equipped with an Intel Core i7-8750H CPU, 16G of RAM and a NVidia Quadro P1000 GPU. Table~\ref{computation-time-kth} and Table~\ref{computation-time-uav} summarize the execution time of each part of our method for the KTH-Action and UAV-Gesture datasets, respectively. For metric notation, $M_1$ refers to Eq.~\eqref{eq:frob-norm-new-metrix} and $M_2$ refers to Eq.~\eqref{eq:distp2}, in Section~\ref{metrics}.



\begin{table}[H]
\centering
\caption{Execution time (in seconds) obtained on the KTH-Action dataset for the different steps of the method for one sequence.}
\begin{adjustbox}{width=0.478\textwidth}
\begin{tabular}{|c|c|c|c|c|}
\hline
\textbf{Pose extraction} & \textbf{Curve Fitting} & \textbf{Alignment method - Metric} & \textbf{Alignment} & \textbf{Classification} \\ \hline
147 & 0.069 & DTW - $M_1$ & 0.034 & 0.41 \\
147 & 0.069 & DTW - $M_2$ & \textbf{0.02} & 0.41 \\ \hline
147 & 0.069 & GAK - $M_1$ & 0.04 & 0.49 \\
147 & 0.069 & GAK - $M_2$ & \textbf{0.019} & 0.49 \\ \hline
\end{tabular}
\end{adjustbox}
\label{computation-time-kth}
\end{table}

\begin{table}[H]
\centering
\caption{Execution time (in seconds) obtained on the UAV-Gesture dataset for the different steps of our method for one sequence.}
\begin{adjustbox}{width=0.478\textwidth}
\begin{tabular}{|c|c|c|c|c|}
\hline
\textbf{Pose extraction} & \textbf{Curve Fitting} & \textbf{Alignment method - Metric} & \textbf{Alignment} & \textbf{Classification} \\ \hline
- & 0.504 & DTW - $M_1$ & 0.128 & 0.038 \\
- & 0.504 & DTW - $M_2$ & \textbf{0.072} & 0.038 \\ \hline
- & 0.504 & GAK - $M_1$ & 0.138 & 0.53 \\
- & 0.504 & GAK - $M_2$ & \textbf{0.066} & 0.053 \\ \hline
\end{tabular}
\end{adjustbox}
\label{computation-time-uav}
\end{table}

For the KTH-Action dataset, we consider a sequence of 61 frames and a sequence of 192 frames for UAV-Gesture. First, we can observe that the pose extraction phase takes most of the execution time for the KTH-Action dataset. This is partially due to the fact that our GPU is not powerful enough (we get around 3.5fps with our Quadro P1000). The extraction time is not reported for the UAV-Gesture dataset as the skeletons are available with the dataset. The second thing we can observe is the low difference in computation time for the alignment part when switching from DTW to GAK. We can also note that when we use $M_2$, the computation time can be reduced by a factor 2 compared to the use of $M_1$, showing that the formula~\eqref{eq:distp2} is in our case cheaper to evaluate than~\eqref{eq:frob-norm-new-metrix}. If we only consider the execution time for the treatment of the skeletons, it takes around 0.499 seconds to classify an action of the KTH-Action dataset and around 0.614 seconds for an action of the UAV-Gesture dataset in the best case scenario.

\section{Conclusion and Future Work}\label{sect:conclusion}
In this paper, we have proposed a method for comparing and classifying temporal sequences of 2D/3D landmarks on the positive semi-definite manifold. Our approach involves three different steps: 1) We build a trajectory on the Riemannian manifold from the body skeletons; 2) we apply a curve fitting algorithm on the trajectories to denoise the data points; 3) we perform a temporal alignment using a Global Alignment Kernel. Our experiments on three publicly available datasets show that the proposed approach gives competitive results with respect to state-of-the-art methods.

\section{Acknowledgements}
We thank Dr. A. Kacem for fruitful discussions. We also thank Prof. J-C. Alvarez Paiva  for fruitful discussions on the formulation of the distance between $n \times 2$ landmarks configurations.

\section{Appendix}
\begin{proof} of Theorem~\ref{theo1}.
We can reformulate our metric introduced in Eq.~\eqref{eq:frob-norm-new-metrix} with:
\begin{align*}
d^2(G_i, G_j) & = \Tr \left[ (Z_jQ - Z_i)(Z_jQ - Z_i)^T \right] \\
  & = \Tr(G_i) - 2\Tr(Z_iQ^TZ_j^T) + \Tr(G_j) .
\end{align*}

To minimize our distance, we need to maximize the term $\Tr(Z_iQ^TZ_j^T)$. Let $Z_j^TZ_i$ be a $2 \times 2$ matrix with four unknown values $a, b, c, d$ and let $Q \in \mathcal{O}_p$, we  maximize:
\begin{equation}
\label{eq:maximize-term}
\begin{split}
    \max\;\Tr \left[ \left( \begin{array}{cc} a\cos{\Theta} - b\sin{\Theta} & - \\ - & c\sin{\Theta} + d\cos{\Theta} \end{array} \right) \right] .
\end{split}
\end{equation}

From Eq.~\eqref{eq:maximize-term} we now have to find the maximum of $(a + d)\cos{\Theta} + (c - b)\sin{\Theta}$, meaning that we have to maximize $\sqrt{(a + d)^2 + (c - b)^2}\cos{(O-O')}$. As we want to maximize this value, $O$ has to be equal to $O'$, so $\sqrt{(a + d)^2 + (c - b)^2}\cos{(O-O')} \leqslant \sqrt{(a + d)^2 + (c - b)^2}$. Therefore we can say that:
\begin{equation}
    \max \Tr(Z_iQ^TZ_j^T) = \sqrt{(a + d)^2 + (c - b)^2} .
\end{equation}
\end{proof}



{\small
\bibliographystyle{ieee}
\bibliography{egbib}
}

\end{document}